\PassOptionsToPackage{numbers,compress,sort}{natbib}
\documentclass{article}

\usepackage[main, final]{neurips_2026}

\usepackage{import}
\usepackage[utf8]{inputenc}
\usepackage[T1]{fontenc}
\usepackage{hyperref}
\usepackage{xurl}
\usepackage{graphicx}
\usepackage{booktabs}
\usepackage{amsmath,amssymb,amsfonts}
\usepackage[table]{xcolor}
\usepackage{float}
\usepackage{algorithm}
\usepackage{algorithmic}
\usepackage{multirow}
\usepackage{subcaption}
\usepackage{wrapfig}
\usepackage{placeins}
\usepackage{enumitem}
\usepackage{nicefrac}
\usepackage{silence}
\usepackage{microtype}
\usepackage{marvosym}
\usepackage{pifont}
\usepackage[capitalise,noabbrev]{cleveref}
\usepackage{xspace}

\newcommand{\framework}{ParaVT}
\newcommand{\method}{PARA-GRPO}

\newcommand{\ie}{\emph{i.e.,}\xspace}
\newcommand{\eg}{\emph{e.g.,}\xspace}

\makeatletter
\newcommand{\blfootnote}[1]{%
  \begingroup
  \renewcommand{\thefootnote}{}%
  \def\@thefnmark{}%
  \begin{NoHyper}\footnotetext{#1}\end{NoHyper}%
  \endgroup
}
\makeatother

\newsavebox{\envbox}
\sbox{\envbox}{\raisebox{0.9ex}{\scalebox{0.6}{\Letter}}}
\newcommand{\envmark}{\usebox{\envbox}}

\title{\framework{}: Taming the Tool Prior Paradox for Parallel Tool Use in Agentic Video Reinforcement Learning}

\author{%
Zuhao Yang\textsuperscript{1,2,7} \enspace
Kaichen Zhang\textsuperscript{1,3,7} \enspace
Sudong Wang\textsuperscript{1,4} \enspace
Keming Wu\textsuperscript{1,5,7} \enspace
Zhongyu Yang\textsuperscript{6} \\[2pt]
\bfseries Bo Li\textsuperscript{7} \enspace
Xiaojuan Qi\textsuperscript{3} \enspace
Shijian Lu\textsuperscript{2,}\envmark \enspace
Xingxuan Li\textsuperscript{1,}\envmark \enspace
Lidong Bing\textsuperscript{1} \\[6pt]
{\normalfont
\textsuperscript{1}Apodex \quad
\textsuperscript{2}NTU \quad
\textsuperscript{3}HKU \quad
\textsuperscript{4}HKUST(GZ) \quad
\textsuperscript{5}THU \quad
\textsuperscript{6}LZU \quad
\textsuperscript{7}LMMs-Lab} \\[4pt]
{\normalfont\small\url{https://evolvinglmms-lab.github.io/ParaVT/}}%
}

\begin{document}

\maketitle

\blfootnote{Work done during Z.Y., K.Z., S.W., and K.W.'s internships at Apodex. This work was supported by Apodex, which provided the compute, storage, and engineering infrastructure for the reported experiments.\envmark\ Corresponding Authors.}

\begin{abstract}
Training large multimodal models (LMMs) via reinforcement learning (RL) to natively invoke video-processing tools (\eg cropping) has become a promising route to long-video understanding.
However, most existing agentic Video-LMMs dispatch tool calls sequentially (\ie one per turn).
Sequential tool calling can propagate localization errors, accumulate irrelevant context, and require an additional main-model generation for each successive crop.
We introduce \textbf{\framework{}}, a multi-agent RL framework for \textbf{Para}llel \textbf{V}ideo \textbf{T}ool calling, dispatching multiple time-window crops in a single turn, allowing evidence from multiple temporal windows to be examined concurrently.
Yet applying standard Group Relative Policy Optimization (GRPO) to \framework{} reveals an obstacle we term the \emph{Tool Prior Paradox}: inherited tool-use priors appear to support exploration while competing with the output format learned during supervised fine-tuning. Standard RL can also settle on tool skipping.
A comparison with a model with weaker tool-use priors shows stable formatting but no tool calls during RL, consistent with a trade-off between format stability and tool exploration.
We propose \textbf{\method{}} (\textbf{P}arseability-\textbf{A}nchored and \textbf{R}atio-g\textbf{A}ted \textbf{GRPO}), which augments standard RL with two complementary mechanisms: (i) a \emph{format reward} that encourages tag closure and penalizes unclosed reasoning blocks, and (ii) a \emph{per-prompt randomization of the overview-frame budget} to create a reward advantage for tool use when the overview provides insufficient evidence.
Across six long-video understanding benchmarks, \framework{} improves over the Qwen3-VL baseline by $+7.9\%$ on average, with \method{} raising training-time format reward from $0.13$ to a peak of $0.64$.
These findings highlight the need to account for inherited tool-use priors when designing RL methods for long-video understanding.

\end{abstract}

\section{Introduction}
\label{sec:intro}

Long-video understanding spans question answering and temporal grounding~\citep{yang2025timeexpert}, with recent work adopting an \emph{agentic video reasoning} paradigm.
To answer ``Which player took the decisive volley in this ninety-minute soccer match?'', a large multimodal model (LMM) is post-trained to invoke video-processing tools via supervised fine-tuning (SFT) on customized tool-use traces followed by reinforcement learning (RL) with verifiable rewards~\citep{yang2026longvt,zhang2025thinking,ouyang2025conan,ding2025videozoomer,shen2025zoom,jain2025sage,zeng2026videoo3}.
For example, LongVT~\citep{yang2026longvt} trains interleaved multimodal chains of tool-thought through cold-start SFT, agentic RL, and agentic reinforcement fine-tuning, instilling a global-to-local loop that skims the match and zooms into the few seconds of evidence until the answer is grounded.
These methods, however, dispatch tool calls sequentially across turns (\ie one tool call per turn), with successive tool outputs accumulating in a single context window. This paradigm is brittle along three dimensions (\Cref{fig:architecture}a). First, a single mis-localized crop propagates errors with no peer to correct it. Second, multi-turn accumulation aggregates context corruption, and long-context models use information in long inputs unevenly~\citep{liu2024lost}. Third, each crop requires another main-model generation over the accumulated context, so inference cost grows with the number of turns.

\begin{figure}[t]
\centering
\includegraphics[width=0.9\linewidth]{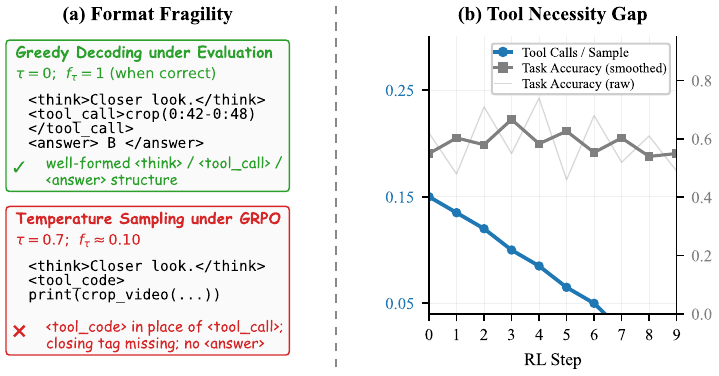}
\caption{\textbf{Two Failure Modes of the Tool Prior Paradox.} \emph{(a) Format Fragility}: rollouts are well-formed under greedy decoding (sampling temperature $\tau{=}0$, format reward $\approx 1$). Under temperature sampling within vanilla GRPO ($\tau{=}0.7$), the policy reverts to the alternative \texttt{<tool\_code>} tag in place of \texttt{<tool\_call>}, often drops closing tags, and stops emitting \texttt{<answer>} altogether (mean format reward $f_\tau{\approx}0.1$). \emph{(b) Tool Necessity Gap}: tool-call count drops to near-zero within $7$ steps while task accuracy oscillates between $0.45$ and $0.74$, as the policy converges on the shortcut of skipping tools.}
\label{fig:obstacles}
\end{figure}

To this end, we introduce \textbf{\framework{}}, a multi-agent RL framework for \textbf{Para}llel \textbf{V}ideo \textbf{T}ool calling (\Cref{fig:architecture}b).
Within \framework{}, a main agent issues multiple temporal-window crops in a single turn and aggregates the returned evidence for decision-making.
During RL rollouts, these crops are processed in parallel by sub-agents that share the policy's weights, each returning a text summary of its assigned window.
The main agent can cross-check these summaries, reducing reliance on any single crop.

\begin{wrapfigure}[19]{r}{0.5\linewidth}
\centering
\includegraphics[width=\linewidth]{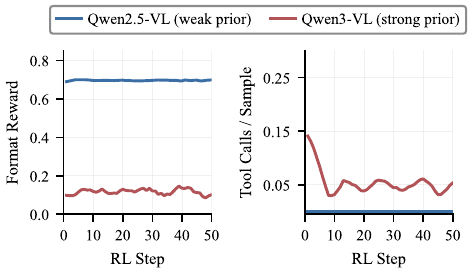}
\caption{\textbf{Cross-Model Evidence for the Tool Prior Paradox under Vanilla GRPO.} Qwen3-VL-8B (stronger tool prior) explores tool use but collapses on format, while Qwen2.5-VL-7B (weaker tool prior) stays format-perfect yet emits zero tool calls.}
\label{fig:cross_model}
\end{wrapfigure}

A natural choice for end-to-end \framework{} training is Group Relative Policy Optimization (GRPO)~\citep{guo2025deepseek} on top of a tool-native cold-started Qwen3-VL~\citep{qwen2025qwen3vl} checkpoint. However, vanilla GRPO exhibits two coupled training-time failures.
The first is \emph{Format Fragility} (\Cref{fig:obstacles}a): the SFT-learned \texttt{<think>}/\texttt{<tool\_call>}/\texttt{<answer>} format is reliable under greedy decoding but, within a few vanilla-GRPO steps under temperature sampling, the policy reverts to the alternative \texttt{<tool\_code>} schema. This is a shallow override of the SFT format reminiscent of the Superficial Alignment Hypothesis~\citep{zhou2023lima}, compounded by the competing \emph{inherited tool-use priors}: the probability mass on tool-call continuations already present in the base checkpoint before our SFT stage that resurfaces under RL-time temperature. As a result, malformed rollouts cannot be parsed into rewardable tool calls, so the GRPO advantage signal is computed over a corrupted trajectory population before any tool-use credit can be assigned.
The second is \emph{Tool Necessity Gap} (\Cref{fig:obstacles}b): when uniformly sampled overview frames suffice to answer many prompts directly, rewards need not favor tool-calling over tool-skipping rollouts. GRPO then provides little consistent task-reward incentive for tool use, and the policy can converge to the shortcut of skipping tools~\citep{skalse2022reward}.

To probe the role of inherited tool-use priors, we replicate the same setup on Qwen2.5-VL~\citep{qwen2025qwen25vl} (with much weaker tool priors than Qwen3-VL) under identical hyperparameters (\Cref{fig:cross_model}): its format stays near-perfect, yet RL elicits no tool calls.
This cross-model contrast suggests a trade-off: stronger tool-use priors may support exploration while competing with the cold-started output format, whereas the weaker-prior model retains its format but does not explore tools. The comparison does not isolate prior strength from other model differences. We term this observed tension the \emph{Tool Prior Paradox}.
This brings us to the central question of this work: \emph{for tool-native LMMs, does the inherited tool-use prior help or hurt tool use after RL?}

We propose \textbf{\method{}} (\textbf{P}arseability-\textbf{A}nchored and \textbf{R}atio-g\textbf{A}ted \textbf{GRPO}) with \emph{Exploration Anchoring} and \emph{nFrames Gating} to tame the Tool Prior Paradox (\Cref{sec:paragrpo}).
Exploration Anchoring combines a format reward that encourages tag closure and penalizes unclosed reasoning blocks with a Constrained Generation hook that fixes the opening reasoning tag. Together they anchor rollout parseability without restricting reasoning content or tool-call sequences. Both mechanisms adapt practices popularized by DeepSeek-R1~\citep{guo2025deepseek}: its \texttt{<think>} reasoning template and rule-based format rewards. We adopted this variant after a standard format reward alone failed to keep rollouts parseable.
nFrames Gating tackles the reward-signal side: randomizing the overview-frame budget per prompt makes some prompts harder to answer without tools, creating opportunities for a within-group reward contrast between tool-calling and tool-skipping rollouts.
The two design choices are complementary. Anchoring preserves parseability so that tool-use credit can be assigned, while gating changes the evidence available before a tool call. The combined recipe achieves higher format reward than either single-component variant while retaining tool use (\Cref{sec:ablation}). Empirically, \method{} raises training-time format reward from $0.13$ to a peak of $0.64$ and improves the agentic-setting Qwen3-VL baseline on every tested benchmark (\Cref{sec:main_results}).

Our contributions are fourfold.
\textbf{(i)} We introduce \framework{}, a framework for parallel video-tool calling via agentic RL, trained on a self-curated $97$K-sample multi-task SFT corpus and a separate $4.4$K-sample RL corpus covering video question answering (QA) and temporal grounding. \emph{Code, data, and model weights are publicly available.}
\textbf{(ii)} We identify the Tool Prior Paradox, decompose it into Format Fragility and Tool Necessity Gap, and support the diagnosis with a cross-model contrast on a weak-prior LMM.
\textbf{(iii)} We propose \method{}, which composes nFrames Gating, a per-prompt randomization of the overview-frame budget that restores a reward contrast between calling and skipping the tool, with Exploration Anchoring to tackle the Tool Necessity Gap and Format Fragility together.
\textbf{(iv)} We conduct extensive comparisons with existing methods on six long-video benchmarks and systematic ablations of \method{}'s key design choices, demonstrating the effectiveness of \framework{}.

\section{Related Work}

\paragraph{Long-Video Understanding.}
RL-based long-video understanding spans three broad directions: (i) \emph{tool-free RL}~\citep{feng2025video,wang2025videorft,li2025videochat,wang2025videothinker,wang2025time,zhang2025rewatchr1} optimizes textual reasoning without external tool calls; (ii) \emph{multi-agent and orchestrated RL} jointly optimizes cooperating policy agents~\citep{chen2025videochatm1} or trains a planner to query specialized perception agents~\citep{liu2025longvideoagent,tian2025egor1}; (iii) \emph{single-LMM adaptive perception and tool-augmented RL} revisits visual evidence through additional sampling or tool calls: LongVT~\citep{yang2026longvt} (sequential \texttt{crop\_video} calls), VITAL~\citep{zhang2025thinking} (on-demand frame sampling over multiple rounds), Zoom-Zero~\citep{shen2025zoom} (a single coarse-to-fine zoom-in pass), Conan~\citep{ouyang2025conan} (an identify-reason-act loop over frames), VideoZoomer~\citep{ding2025videozoomer} (iterative \texttt{<video\_zoom>} calls), LoVe-R1~\citep{fu2025lover1} (step-decoupled iterative zoom-in), SAGE~\citep{jain2025sage} (a JSON tool-action schema), VideoChat-R1.5~\citep{yan2025videochatr15} (iterative re-perception of predicted regions), and Video-o3~\citep{zeng2026videoo3} (multi-hop clue seeking), together with methods that select segments or frames over successive turns~\citep{xie2025videomtr,he2025framethinker,pan2025timesearchr}. Our work differs in two respects: (1) we train a tool-native LMM to dispatch several temporal crops per turn, combining serial context expansions; (2) we identify and address the Tool Prior Paradox, a failure mode of tool-native RL that, to our knowledge, prior work has not characterized.

\paragraph{Parallel and Multi-Agent Tool Use.}
Executing independent tool calls concurrently is established for text agents, from compiler-style parallel function calling~\citep{kim2024llmcompiler} to RL-trained parallel search queries and reasoning threads~\citep{zhao2025parallelsearch,pan2025apr}. With RL, MATPO~\citep{mo2025matpo} trains planner and worker roles inside one large language model (LLM) through role-specific prompts. WideSeek-R1~\citep{xu2026wideseekr1} jointly optimizes a lead agent and parallel sub-agents that share one LLM, and Kimi K2.5~\citep{kimiteam2026kimik25} trains an orchestrator whose parallel sub-agents are frozen. For video, systems built on off-the-shelf models process clips or frame subsets in parallel~\citep{pang2025mrvideo,chung2025vps,eltahir2026videoatlas} or coordinate several LMM agents~\citep{chen2025lvagent,yang2026svagent}. MACF~\citep{chen2026macf} trains segment-level agents with a supervised curriculum. Concurrent with our work, ReTool-Video~\citep{liu2026retoolvideo} trains a planner with GRPO that may issue several independent actions per round over a large tool library. \framework{} trains parallel temporal-crop dispatch in a tool-native LMM, with sub-agents sharing the current policy's weights but excluded from the RL loss.

\paragraph{Format Stability and Tool Use in RL.}
In agentic RL, format stability is a precondition for tool-use learning: only parseable rollouts can be credited for their tool calls.
Rule-based format rewards, popularized by DeepSeek-R1~\citep{guo2025deepseek}, are standard in open recipes, either as per-tag counts~\citep{openr12025} or as structural terms combined with semantic rewards, where format collapse under semantic pressure has also been reported~\citep{benchaliah2025datarus}. More generally, constrained decoding restricts generation to a grammar~\citep{willard2023efficient}.
The Superficial Alignment Hypothesis characterizes post-training as learning style and format~\citep{zhou2023lima}, though this hypothesis remains contested~\citep{raghavendra2024revisiting}. In safety alignment, distributional changes can concentrate in the first few output tokens~\citep{qi2024safety}. Our Format Fragility is analogous but specific to tool-native LMMs under RL-time temperature sampling: the SFT-learned \texttt{<tool\_call>} tag reverts to the alternative \texttt{<tool\_code>} tag under RL rollouts, fragmenting the structural-boundary distribution. PRISM inserts on-policy distillation between SFT and RL, using mixture-of-experts perception and reasoning feedback to reduce distributional drift~\citep{wang2026prism}. \framework{} instead intervenes during RL, leaving the SFT-to-RL handoff unchanged.
At the token level, RL-induced policy shifts concentrate on a sparse subset of high-divergence tokens~\citep{meng2026sparse}. To encourage exploration on tokens that drive correct outcomes, prior work relaxes the Kullback--Leibler penalty on those tokens~\citep{vassoyan2025ignore}. Exploration Anchoring differs from these gradient-side interventions: it fixes the opening reasoning tag and adds a sequence-level reward that checks only structural boundary tags.
Related efforts shape when a tool is worth calling. AdaTooler-V~\citep{wang2025adatoolerv} rescales rewards by an offline estimate of each sample's tool benefit. Pixel Reasoner~\citep{wang2025pixelreasoner} adds a curiosity bonus against the reluctance to adopt new visual operations. FrameMind~\citep{ge2025framemind} assigns each rollout in a group a different frame-count and resolution configuration. nFrames Gating instead randomizes the overview-frame budget per prompt, so that the call/skip reward contrast arises from the training distribution.

\section{Method}
\suppressfloats[t]
\label{sec:method}

\subsection{ParaVT: Parallel Video Tool Calling for Long-Video Understanding}
\label{sec:arch}

\begin{figure}[t]
\centering
\includegraphics[width=0.9\linewidth]{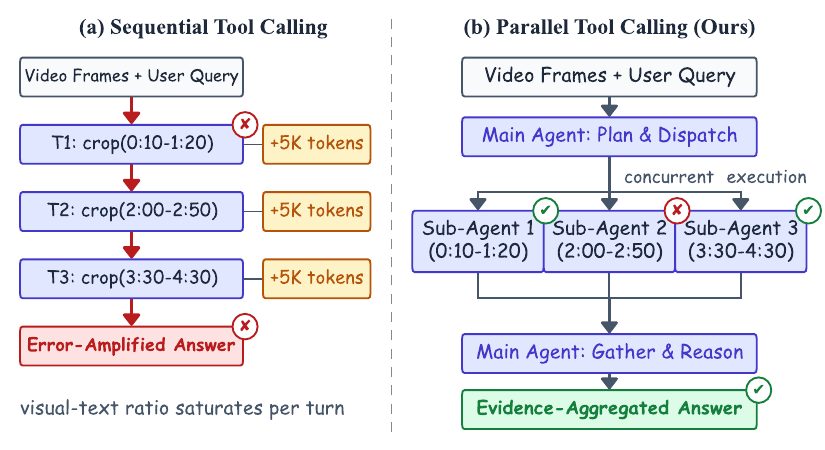}
\caption{\textbf{Framework Comparison.} \emph{(a) Sequential Tool Calling}: successive turns re-include the full context, accumulating visual-token overhead. A single mis-localized crop (\textcolor[HTML]{C44A3F}{\ding{55}}) propagates errors with no peer to correct it, yielding an error-amplified answer. \emph{(b) Parallel Tool Calling (Ours)}: one main agent dispatches $K$ tool calls concurrently to $K$ independent sub-agents (shown for $K{=}3$). Mis-localized peers (\textcolor[HTML]{C44A3F}{\ding{55}}) can be cross-checked against correct ones (\textcolor[HTML]{5FA478}{\ding{51}}), yielding an evidence-aggregated answer.}
\label{fig:architecture}
\end{figure}

\framework{} consists of three design elements: a parallel-dispatch architecture (\Cref{sec:framework}), a two-stage training pipeline (\Cref{sec:training}), and a self-curated multi-task dataset (\Cref{sec:data}).

\subsubsection{Framework Design}
\label{sec:framework}

A common paradigm for tool-augmented long-video understanding lets the LMM decide \emph{when} and \emph{where} in the video to look more closely by issuing a \texttt{crop\_video(start, end)} function call that returns the requested temporal segment with densely resampled frames for further inspection (\Cref{fig:architecture}a).
Existing realizations of this design~\citep{yang2026longvt,zhang2025thinking,ouyang2025conan,ding2025videozoomer} request additional visual evidence \emph{sequentially}: one tool action per turn, with the returned frames re-injected into the running context before the next turn begins.

\framework{} re-organizes the same loop as a single-turn divide-and-conquer step (\Cref{fig:architecture}b). Within a single turn, the main agent emits $K$ \emph{parallel} \texttt{<tool\_call>} invocations on selected temporal windows, each dispatched to one of $K$ independent sub-agents that share weights with the main agent, a multi-agent-in-one-model design in which roles differ only by prompt~\citep{mo2025matpo}. In RL rollouts, a sub-agent is a fresh single-turn call of the current policy under a fixed, query-agnostic summarization prompt that sees only the frames of its assigned window.
Each sub-agent processes only its assigned window and returns a textual summary rather than resampled frames. The $K$ calls run concurrently. Each summary is truncated to a fixed length and appended to the main agent's context as a \texttt{<tool\_response>} block, on which the main agent reasons to generate the final \texttt{<answer>}.

This single-turn parallel dispatch yields three concrete advantages over the sequential paradigm.
\emph{(i) Peer-Correctable Evidence.} The main agent cross-checks $K$ independent window summaries, allowing peers to correct a mis-localized crop before the error propagates.
\emph{(ii) Controlled Context Growth.} All $K$ crops of a turn enter the context at once. The accumulated context is therefore re-encoded once per turn rather than once per crop. In RL rollouts, text summaries further limit the extension to a small constant per crop, while returning frames would add $K$ visual-token blocks per turn.
\emph{(iii) Bounded Inference Latency.} Issuing the $K$ crops in one turn replaces $K$ sequential main-model generations with one. During RL rollouts, the $K$ sub-agents run concurrently, so the tool-using portion is bounded by the slowest sub-agent rather than by their sum.

\subsubsection{Training Strategy}
\label{sec:training}

\paragraph{Cold-Start SFT with Parallel Tool Traces.}
The base LMM (\ie Qwen3-VL-8B-Instruct~\citep{qwen2025qwen3vl}) supports the \texttt{<tool\_call>} schema, but it is not trained on the single-turn $K$-crop format that \framework{} requires.
Without supervised exposure to parallel traces, probe RL runs from the base checkpoint fail to produce parseable rollouts (\Cref{app:rollouts}), and the GRPO advantage signal collapses before any tool-use credit can be assigned, consistent with the observation that native tool-calling abilities are too weak for direct RL training~\citep{yang2026longvt}.
Therefore, we conduct an SFT cold start on the base model with the parallel-tool corpus and select an early checkpoint as the RL initialization based on training-time format stability under temperature sampling. The two-stage SFT-then-RL pipeline is a common recipe for open multimodal-reasoning systems~\citep{huang2026vision,zhang2026openmmreasoner}. \framework{} specializes it to parallel video-tool calling with the corpus described in \Cref{sec:data} and the reward design in \Cref{sec:paragrpo}.

\paragraph{Agentic RL with Verifiable Rewards.}
Starting from the cold-started checkpoint, we conduct GRPO with two verifiable reward terms: an accuracy term against the ground-truth answer and a format term over the \texttt{<think>}/\texttt{<tool\_call>}/\texttt{<answer>} schema. For each prompt, GRPO samples $G{=}8$ rollouts and updates the policy by their group-normalized advantage. Rewards and advantages are computed on main-agent tokens only, and sub-agent generations are excluded from the loss, so RL optimizes when to dispatch and how to aggregate, while sub-agents change only through the shared weights. Vanilla GRPO at this stage exposes the Format Fragility and Tool Necessity Gap introduced in \Cref{sec:intro}.
\method{} addresses these failures (\Cref{sec:paragrpo}).

\subsubsection{Data Curation}
\label{sec:data}

\paragraph{SFT Split.}
The SFT corpus contains $97$K samples spanning four task families (full per-source breakdown in \Cref{tab:sft_data}): general video QA ($50$K from LLaVA-Video-178K~\citep{zhang2024llavavideo}), long-video reasoning chains ($5$K from LongVideo-Reason~\citep{chen2025longrl}), temporal grounding ($12$K Charades-STA~\citep{gao2017charades} direct grounding + $6$K Charades-STA-converted traces with parallel tool calls), and self-curated $22.5$K parallel-tool traces.
The mix preserves general video understanding while giving the model concentrated supervision on the parallel multi-tool schema. Tool-using samples account for $30\%$ of the corpus. We chose this fraction after an earlier larger mix ($212$K total at $14\%$ tool) yielded weaker downstream tool-calling than this smaller, tool-richer mix (\Cref{app:tool_fraction}).

The parallel-tool traces are drawn from three sources: $15$K LongVT~\citep{yang2026longvt} tool-using rollouts, $5$K Gemini-2.5-Flash~\citep{team2025gemini25} distillations of LongVT prompts, and $2.5$K multi-segment grounding samples from MUSEG~\citep{luo2025museg}.
The first two sources emit one \texttt{crop\_video} call per assistant turn with resampled video frames re-injected into the next turn's context, a \emph{sequential} format that does not exhibit the single-turn $K$-call schema we want \framework{} to learn.
We traverse each sequential trace and merge adjacent crops whose target windows do not overlap and whose tool responses do not cross-reference each other (\eg ``inspect 00:30--00:50'' followed by ``inspect 02:10--02:25'' on independent visual evidence). Calls that fail this independence check, such as a refinement crop conditioning on its predecessor, remain sequential.
Each tool's visual response is then replaced by a textual summary of the segment, aligning the SFT data with the RL sub-agent's text-summary output format and keeping context length manageable when several crops appear in the same response.

\paragraph{RL Split.}
The RL corpus aggregates $4{,}406$ samples from three task families: $1{,}606$ open-ended QA from filtered LongVT~\citep{yang2026longvt} RL data, $1{,}600$ multiple-choice questions (MCQ) from the VideoR1~\citep{feng2025video} RL pool, and $1{,}200$ temporal video grounding (TVG) queries from the Charades-STA~\citep{gao2017charades} training set. The SFT and RL corpora share some Charades training videos.
Before training begins, we apply an offline difficulty filter inspired by DAPO~\citep{yu2026dapo}: we exclude open-ended prompts whose ground-truth answers exceed $15$ words and prompts with unanimously unsuccessful rollouts under the cold-started policy. These are heuristic exclusions, rather than a guarantee of zero advantage under future policies.

\subsection{\method{}: Parseability-Anchored and Ratio-Gated GRPO}
\label{sec:paragrpo}

Vanilla GRPO faces two obstacles to learning tool use in \framework{}. \emph{Format Fragility} manifests as tool-tag reversion or unclosed structural tags. Malformed rollouts can disrupt tool execution and the assignment of tool-use credit. The \emph{Tool Necessity Gap} persists even on parseable rollouts: when overview frames already suffice to answer a prompt, task rewards need not consistently favor tool-calling over tool-skipping rollouts.

\method{} addresses these obstacles with two complementary components. \emph{Exploration Anchoring} stabilizes structural boundaries through constrained generation and rewards for tag closure. \emph{nFrames Gating} randomizes the per-prompt overview-frame budget so that some GRPO groups contain a reward contrast between tool-calling and tool-skipping rollouts. Anchoring preserves parseability, while gating supplies an incentive to use tools when overview frames are insufficient. Increasing temperature can encourage tool exploration but destabilize tag closure. Temperature adjustment does not replace the reward contrast that nFrames Gating is designed to create.

\subsubsection{Exploration Anchoring}

Changing the SFT target tag did not resolve Format Fragility in our probe. Because the main runs revert from \texttt{<tool\_call>} to \texttt{<tool\_code>}, we also tested SFT directly on \texttt{<tool\_code>}. The resulting policy still emits \texttt{<tool\_call>} more often than its SFT-trained tag during RL (\Cref{app:bidirectional}). This bidirectional reversion indicates that choosing the alternative tag does not eliminate the competing tool representations. We therefore retain the native \texttt{<tool\_call>} tag and target structural-boundary stability during RL.

Exploration Anchoring uses two mechanisms to restore the closing \texttt{</think>} and \texttt{</answer>} tags where structural collapse concentrates.

\paragraph{Constrained Generation.}
At the entry and exit of the response, two minimal interventions reinforce what SFT has already taught reliably. A \emph{Think Prefix} pins the first tokens of every response to \texttt{<think>$\backslash$n}, ruling out blind direct answers and tool calls without restricting what the model reasons about. It follows the \texttt{<think>} template convention of DeepSeek-R1~\citep{guo2025deepseek}. A complementary \emph{Answer Suffix} term in the format reward credits the presence of a final \texttt{<answer>} block even when intermediate structure is imperfect, so policies that recover into a well-formed answer are not penalized for exploration along the way.

\paragraph{Selective Anchoring.}
To address failures to close structural tags, we add an anchoring term to the format reward:
\begin{equation}
\label{eq:anchor}
\begin{aligned}
R_{\text{anchor}}(y) ={}& \alpha\,\mathbf{1}\bigl[\text{reasoning block closed}\bigr] - \gamma\,\mathbf{1}\bigl[\texttt{<think>}\text{ opened but never closed}\bigr] \\
&+ \beta\,\mathbf{1}\bigl[\text{full }\texttt{<think>}\!\to\!\texttt{</think>}\!\to\!\texttt{<answer>}\text{ flow}\bigr],
\end{aligned}
\end{equation}
where $R_{\text{anchor}}$ is a sequence-level scalar computed from string checks on the completed rollout $y$. It contributes to the total reward, whose group-normalized advantage GRPO broadcasts to all trained main-agent tokens, changing rollout preference rather than assigning token-level credit. Unlike equal-credit tag-count rewards~\citep{openr12025}, the anchor penalizes an opened but unclosed \texttt{<think>}.
The triplet $(\alpha, \beta, \gamma)$ and the outer scaling $\lambda_{\text{anchor}}$ inside $R_{\text{fmt}} = R_{\text{base}} + \lambda_{\text{anchor}} R_{\text{anchor}}$ govern how aggressively the anchor pulls the policy toward parseability.

Constrained Generation and Selective Anchoring address different parts of the response: the former fixes the entry and credits the final answer, the latter rewards closure of internal boundaries. Neither restricts the reasoning or tool-call content that lives between them.

\subsubsection{nFrames Gating}

Anchoring restores parseable rollouts, but a reward advantage for tool use is still needed. With a generous overview budget, both tool-calling and tool-skipping rollouts may reach the correct answer. When task rewards do not systematically favor tool-calling rollouts within a prompt group, GRPO provides no consistent task-reward incentive for tool use.

nFrames Gating creates the missing gap by randomizing the overview-frame budget per prompt:
\begin{equation}
\label{eq:gating}
n \sim \mathrm{Uniform}\bigl(\{4, 8, 16, 32, 64\}\bigr),
\end{equation}
where $n$ is the number of overview frames seen by all $G{=}8$ rollouts in the GRPO group for that prompt. Reduced budgets ($n{<}64$) push part of the visual evidence outside the overview. Rollouts that recover that evidence through \texttt{crop\_video} can then receive higher task rewards than those relying on the truncated overview. The largest budget ($n{=}64$) preserves the easy regime in which direct answering is sufficient when warranted. Each training step therefore samples a mixture of budget-bound and budget-free prompts, so a controllable fraction of prompts exhibits a non-trivial reward contrast between tool-calling and tool-skipping rollouts, while on prompts where the full budget already suffices, the policy is free to skip tools without penalty. Setting this fraction too low leaves the gating signal too sparse for GRPO to learn from. Setting it too high crowds out the easy regime and induces over-calling.

\subsubsection{Reward Modeling}

Let $x$ denote a prompt (user query paired with the video input), $y$ a rollout, and $a^*$ the ground-truth answer. The composite reward sums three terms:
\begin{equation}
\label{eq:reward}
R(x, y) = R_{\text{acc}}(y, a^*) + \lambda_{\text{fmt}} \, R_{\text{fmt}}(y) + R_{\text{tool}}(y).
\end{equation}
$R_{\text{acc}}$ scores the rollout against the ground truth using a task-appropriate metric (exact match for MCQ, temporal intersection over union (IoU) for grounding, token-level F$1$ for open-ended QA). $R_{\text{fmt}}$ scores structural compliance and embeds $R_{\text{anchor}}$ from \Cref{eq:anchor}, including the penalty for unclosed tags. $R_{\text{tool}}$ rewards well-formed tool invocations.

\section{Experiments}
\label{sec:exp}

\subsection{Implementation Details}
\label{sec:setup}

\paragraph{Training.}
We initialize from Qwen3-VL-8B-Instruct~\citep{qwen2025qwen3vl} and SFT-cold-start on a $97$K multi-task corpus. An early checkpoint is selected for RL initialization based on training-time format stability (selection details in \Cref{app:impl}). RL uses a separately constructed $4{,}406$-sample set after offline difficulty filtering inspired by DAPO~\citep{yu2026dapo}.
We sample $G{=}8$ rollouts at $\tau{=}0.7$, set the anchor weight to $\lambda_{\text{anchor}}{=}0.5$, and decode up to $16$ frames per sub-agent crop.
Training leverages AReaL~\citep{fu2026areal} on a node of $8$ NVIDIA GPUs ($80$~GB+ VRAM each), with $7$ allocated to Fully Sharded Data Parallel (FSDP) training and $1$ to SGLang rollout serving.
Full hyperparameters are listed in \Cref{tab:hyperparams} of \Cref{app:impl}.

\paragraph{Evaluation.}
We evaluate on six long-video benchmarks under a unified $64$-frame overview protocol, reporting MCQ accuracy on VideoMME~\citep{fu2025video}, LongVideoBench~\citep{wu2024longvideobench}, LVBench~\citep{wang2025lvbench}, MLVU~\citep{zhou2025mlvu}, and MMVU~\citep{zhao2025mmvu}, and mean intersection over union (mIoU) on Charades-STA~\citep{gao2017charades}.
\Cref{tab:main} groups open-source baselines by their training paradigm into three settings: \emph{direct-answer} for instruct backbones with no native thinking pattern, \emph{reasoning-enhanced} for models optimized for explicit reasoning, and \emph{tool-augmented} for models with tool-use or adaptive re-perception mechanisms. GPT-4o and Gemini 1.5 Pro are included as proprietary reference rows using published benchmark scores.
We select each baseline's prompt class according to its training paradigm. These settings can differ from those used in the original model report.
We restrict our evaluation to natively post-trained single-LMM methods, excluding agent frameworks~\citep{chen2025lvagent,zhang2025deepvideodiscovery,ye2025tstar,liu2025longvideoagent,yang2026svagent} to keep the comparison fair.
These systems combine separately configured perception or reasoning components, so their reported results are not directly matched to our single-policy evaluation protocol.

\subsection{Main Results}
\label{sec:main_results}

\begin{table*}[!ht]
\centering
\caption{\textbf{Performance Comparison with Existing Video-LMMs.} The best result is in \textbf{bold}; \underline{underline} marks \framework{}'s value when it is not the benchmark-wise best. $\ast$ marks cells withheld due to benchmark--training-data overlap. $\dagger$ marks an evaluation whose native tool-call schema could not be reconciled with the Charades-STA grounding-output format under our unified protocol, so the resulting outputs are not measurable with mIoU.}
\label{tab:main}
\resizebox{\textwidth}{!}{%
\begin{tabular}{l | c c c c c c c}
\toprule[1pt]
\textbf{Model} &
\begin{tabular}[c]{@{}c@{}}\textbf{VideoMME}\\w/o sub\end{tabular} &
\begin{tabular}[c]{@{}c@{}}\textbf{VideoMME}\\w/ sub\end{tabular} &
\begin{tabular}[c]{@{}c@{}}\textbf{LongVideo-}\\\textbf{Bench}\end{tabular} &
\textbf{LVBench} & \textbf{MLVU} & \textbf{MMVU} &
\begin{tabular}[c]{@{}c@{}}\textbf{Charades-STA}\\test\end{tabular} \\
\midrule\midrule
\multicolumn{8}{c}{\textbf{\emph{Proprietary Video-LMMs: best-setting numbers from official reports (not under our unified protocol)}}} \\
\midrule
\textcolor{gray}{GPT-4o~\citep{hurst2024gpt4o}} & \textcolor{gray}{71.9} & \textcolor{gray}{77.2} & \textcolor{gray}{66.7} & \textcolor{gray}{34.7} & \textcolor{gray}{64.6} & \textcolor{gray}{66.7} & \textcolor{gray}{\textemdash{}} \\
\textcolor{gray}{Gemini 1.5 Pro~\citep{team2024gemini15}} & \textcolor{gray}{75.0} & \textcolor{gray}{81.3} & \textcolor{gray}{64.4} & \textcolor{gray}{33.1} & \textcolor{gray}{74.3} & \textcolor{gray}{65.8} & \textcolor{gray}{\textemdash{}} \\
\midrule
\multicolumn{8}{c}{\textbf{\emph{Open Instruct Video-LMMs: direct-answer setting (native single-pass prompt)}}} \\
\midrule
Qwen2.5-VL-7B~\citep{qwen2025qwen25vl} & 55.7 & 64.5 & 46.4 & 32.2 & 47.8 & 65.4 & 31.6 \\
Eagle2.5-8B~\citep{chen2025eagle25} & 57.6 & 63.8 & 54.5 & 37.9 & 52.9 & 58.2 & \textemdash{} \\
\midrule
\multicolumn{8}{c}{\textbf{\emph{Open Reasoning Video-LMMs: reasoning-enhanced setting (\texttt{<think>$\rightarrow$<answer>})}}} \\
\midrule
Video-R1-7B~\citep{feng2025video} & 57.6 & 66.0 & 57.4 & 36.9 & 61.6 & 61.3 & 25.4 \\
VideoChat-R1-7B~\citep{li2025videochat} & 50.4 & 58.2 & 49.2 & 23.8 & 58.7 & 65.0 & 31.5 \\
VideoRFT-7B~\citep{wang2025videorft} & 58.5 & 65.6 & 55.1 & 38.0 & 44.9 & 42.7 & 18.7 \\
Time-R1-7B~\citep{wang2025time} & 58.9 & 66.2 & 56.0 & 38.2 & 60.5 & 63.4 & 34.7 \\
ReWatch-R1-7B~\citep{zhang2025rewatchr1} & 58.8 & 65.0 & 53.6 & 38.5 & 60.1 & 59.8 & 20.2 \\
Video-Thinker-7B~\citep{wang2025videothinker} & 61.9 & 65.3 & 56.0 & $\ast$ & \textbf{65.2} & 64.5 & 29.0 \\
\midrule
\multicolumn{8}{c}{\textbf{\emph{Open Agentic Video-LMMs: tool use or adaptive visual re-perception}}} \\
\midrule
Qwen3-VL-8B~\citep{qwen2025qwen3vl} & 59.9 & 68.4 & 52.2 & 33.1 & 58.3 & 68.0 & 49.3 \\
Conan-7B~\citep{ouyang2025conan} & 55.5 & 62.8 & 54.5 & 38.2 & 59.2 & 64.0 & 25.4 \\
LongVT-RFT-7B~\citep{yang2026longvt} & 59.5 & 66.0 & 54.7 & 37.9 & 59.4 & 63.4 & 23.4 \\
SAGE-7B~\citep{jain2025sage} & 44.1 & 52.4 & 37.4 & 31.8 & 49.7 & 55.7 & 28.9 \\
VideoZoomer-7B~\citep{ding2025videozoomer} & 45.3 & 48.3 & 39.6 & 22.9 & 46.2 & 61.6 & $\dagger$ \\
VideoChat-R1.5-7B~\citep{yan2025videochatr15} & 51.5 & 59.3 & 49.3 & 34.3 & 60.7 & 62.3 & \textemdash{} \\
\rowcolor{gray!15} \textbf{\framework{}-8B (Ours)} & \textbf{62.1} & \textbf{69.4} & \textbf{60.4} & \textbf{39.8} & \underline{65.0} & \textbf{68.6} & \textbf{50.1} \\
\bottomrule[1pt]
\end{tabular}%
}
\end{table*}

\Cref{tab:main} compares \framework{} with existing $7$--$8$B models. \framework{} leads the comparable open-source baselines on six of seven evaluation splits and improves on Qwen3-VL-8B across all splits. Relative gains reach $+15.7\%$ on LongVideoBench, $+20.2\%$ on LVBench, and $+11.5\%$ on MLVU, averaging $+7.9\%$ across the seven splits. It reaches $50.1$ mIoU on Charades-STA and exceeds the GPT-4o reference on LVBench and MMVU.

At inference, $94.1\%$ of $11{,}460$ evaluated samples issue multiple crops within one turn (\Cref{tab:tool_inference}). Parallel dispatch also uses fewer prompt tokens and main-model calls than sequential dispatch, with mean latency decreasing from $80.1$ to $48.5$\,s (\Cref{tab:efficiency}).

\begin{table}[!ht]
\centering

\caption{\textbf{Tool Use at Inference.} Greedy decoding with a $64$-frame overview. Tool use and parallel dispatch are the fractions of samples issuing at least one crop and at least two crops in one turn, respectively. Crops/turn is averaged over crop-emitting turns. MMVU uses a $1{,}000$-question inference set.}
\label{tab:tool_inference}
\small
\begin{tabular}{l r c c c}
\toprule
\textbf{Benchmark} & $N$ & \textbf{Tool use} & \textbf{Crops/turn} & \textbf{Parallel} \\
\midrule
VideoMME w/o sub & $2{,}700$ & $100.0\%$ & $3.53$ & $97.7\%$ \\
VideoMME w/ sub & $2{,}700$ & $99.6\%$ & $3.55$ & $96.4\%$ \\
LongVideoBench & $1{,}337$ & $99.9\%$ & $2.76$ & $82.8\%$ \\
LVBench & $1{,}549$ & $99.9\%$ & $2.96$ & $83.8\%$ \\
MLVU & $2{,}174$ & $100.0\%$ & $3.59$ & $98.3\%$ \\
MMVU & $1{,}000$ & $100.0\%$ & $4.22$ & $99.8\%$ \\
\bottomrule
\end{tabular}

\end{table}

\begin{table}[!ht]
\centering

\caption{\textbf{Inference Cost.} Per-query tokens, main-model calls and latency (mean/p95) at $64$-way client concurrency. Peak memory is per device.}
\label{tab:efficiency}
\small
\resizebox{\linewidth}{!}{%
\begin{tabular}{l r r r c r}
\toprule
\textbf{Dispatch} & \textbf{Prompt tok.} & \textbf{Decode tok.} & \textbf{Calls} & \textbf{Latency (s)} & \textbf{Memory (GB)} \\
\midrule
Sequential & $136{,}502$ & $519$ & $4.23$ & $80.1/168.3$ & $185$ \\
Parallel & $77{,}380$ & $267$ & $2.40$ & $48.5/93.8$ & $165$ \\
\bottomrule
\end{tabular}%
}

\end{table}

\FloatBarrier

\subsection{Ablation Studies}
\label{sec:ablation}

\begin{table*}[!ht]
\centering
\caption{\textbf{Ablation Studies.} Each row reports mean training-time format reward $f_{\tau}$ at sampling temperature $\tau{=}0.7$ and mean number of training-time tool calls per rollout $\kappa$. Both are statistics of the RL training stream, not of inference. The best result is in \textbf{bold}; \underline{underline} marks \framework{}'s value when it is not the block-wise best. Rows shaded \colorbox{gray!15}{gray} mark the full recipe. The \method{}-only row applies \method{} to the base model without the cold start. Block~C compares inference-time dispatch modes on the same trained checkpoint. Thus, $f_{\tau}$ and $\kappa$ are identical across its rows and reported as ``-''.}
\label{tab:ablation_main}
\resizebox{\textwidth}{!}{%
\begin{tabular}{l | c c | c c c c c c}
\toprule[1pt]
\textbf{Setting} & $\boldsymbol{f_\tau}$ & $\boldsymbol{\kappa}$ &
\begin{tabular}[c]{@{}c@{}}\textbf{VideoMME}\\w/o sub\end{tabular} &
\begin{tabular}[c]{@{}c@{}}\textbf{VideoMME}\\w/ sub\end{tabular} &
\begin{tabular}[c]{@{}c@{}}\textbf{LongVideo-}\\\textbf{Bench}\end{tabular} &
\textbf{LVBench} & \textbf{MLVU} & \textbf{MMVU} \\
\midrule\midrule
\multicolumn{9}{c}{\textbf{\emph{(A) Training Stage}}} \\
\midrule
Qwen3-VL-8B & 0.03 & 0.45 & 59.9 & 68.4 & 52.2 & 33.1 & 58.3 & 68.0 \\
\; + SFT Cold-Start & 0.13 & 2.50 & 60.7 & 69.0 & 58.0 & 39.1 & 63.7 & 67.7 \\
\; + SFT Cold-Start + GRPO & 0.13 & 0.02 & 62.0 & 68.6 & 57.5 & 39.3 & 64.5 & 67.5 \\
\; + \method{} only (no cold start) & \textemdash{} & \textemdash{} & 62.0 & \textbf{69.4} & 58.6 & 39.3 & 64.7 & 68.5 \\
\rowcolor{gray!15} \; + SFT Cold-Start + \method{} & 0.41 & 0.21 & \textbf{62.1} & \textbf{69.4} & \textbf{60.4} & \textbf{39.8} & \textbf{65.0} & \textbf{68.6} \\
\midrule
\multicolumn{9}{c}{\textbf{\emph{(B) Component Effectiveness}}} \\
\midrule
Qwen3-VL-8B + Cold-Start + GRPO & 0.13 & 0.02 & 62.0 & 68.6 & 57.5 & 39.3 & 64.5 & 67.5 \\
+ Exploration Anchoring & 0.35 & 0.19 & 61.7 & 68.7 & 59.6 & 39.3 & 64.1 & 67.2 \\
+ nFrames Gating & 0.10 & 1.36 & 61.3 & 68.7 & 58.4 & 39.1 & 63.6 & 65.3 \\
\rowcolor{gray!15} Full PARA-GRPO & 0.41 & 0.21 & \textbf{62.1} & \textbf{69.4} & \textbf{60.4} & \textbf{39.8} & \textbf{65.0} & \textbf{68.6} \\
PARA-GRPO $-$ Tool Reward $R_{\text{tool}}$ & 0.33 & 0.04 & 61.9 & 68.5 & 57.2 & 38.7 & 64.3 & 67.4 \\
PARA-GRPO $-$ Penalty Term $\gamma$ & 0.36 & 0.27 & 61.6 & 69.0 & 58.9 & 38.8 & 64.5 & 67.7 \\
\midrule
\multicolumn{9}{c}{\textbf{\emph{(C) Dispatch Mode}}} \\
\midrule
Sequential Tool Calling & - & - & 61.4 & 68.8 & 57.7 & 37.5 & 64.1 & 66.7 \\
\rowcolor{gray!15} Parallel Tool Calling & - & - & \textbf{62.1} & \textbf{69.4} & \textbf{60.4} & \textbf{39.8} & \textbf{65.0} & \textbf{68.6} \\
\bottomrule[1pt]
\end{tabular}%
}
\end{table*}

\begin{figure}[t]
\centering
\includegraphics[width=\linewidth]{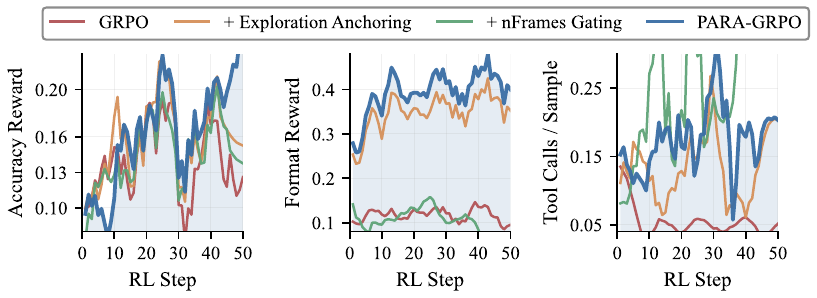}
\caption{\textbf{Training Dynamics across \method{} Components.} Vanilla GRPO (\textcolor[HTML]{B25457}{red}) stays flat at $f_{\tau}{\approx}0.13$ while $\kappa$ collapses to near zero. Exploration Anchoring (\textcolor[HTML]{D6915B}{orange}) lifts $f_{\tau}$ but keeps $\kappa$ moderate. nFrames Gating (\textcolor[HTML]{5FA478}{green}) pushes $\kappa$ off-chart while leaving $f_{\tau}$ low. The full \method{} (\textcolor[HTML]{3A6EA5}{blue}) achieves higher format reward while retaining moderate tool use.}
\label{fig:training_curves}
\end{figure}

\paragraph{Training Stage.}
As shown in Block~A of \Cref{tab:ablation_main}, the cold-started checkpoint over-uses tools ($\kappa{=}2.50$) by directly imitating tool-using demonstrations from SFT traces. Vanilla GRPO swings to the opposite extreme ($\kappa{=}0.02$) by skipping tools within $7$ steps under the reward shortcut (\Cref{fig:obstacles}).
\method{} resolves both extremes, reaching the highest training-time mean format reward in Block~A ($f_\tau{=}0.41$, $\kappa{=}0.21$). It strictly improves on vanilla GRPO across all six evaluation splits, with the largest gains on LongVideoBench and MMVU. The format reward is a shaped score whose scale can change with the ablated reward terms. Separate tag-closure rates are reported in \Cref{tab:format_breakdown}.

\paragraph{Component Effectiveness.}
Block~B shows complementary effects. Exploration Anchoring reaches $(f_\tau,\kappa){=}(0.35,0.19)$, while nFrames Gating reaches $(0.10,1.36)$. The full recipe reaches $(0.41,0.21)$, with higher format reward than either single component and higher accuracy on all six evaluation splits. Removing $R_{\text{tool}}$ reduces $\kappa$ from $0.21$ to $0.04$ and $f_\tau$ from $0.41$ to $0.33$. Removing the unclosed-tag penalty $\gamma$ reduces $f_\tau$ to $0.36$ and costs up to $1.5$~pt on LongVideoBench. Tool utility is not determined by call count. We therefore interpret $\kappa$ jointly with format reward and task accuracy.

\paragraph{Training Dynamics.}
\Cref{fig:training_curves} visualizes the same variant comparison during RL. Vanilla GRPO never recovers either metric: $f_\tau$ stays flat near $0.13$ and $\kappa$ collapses to near zero within $7$ steps, leaving the policy in the format-shortcut regime as introduced in \Cref{sec:intro}. Used in isolation, \emph{Exploration Anchoring} restores format ($f_\tau{\to}0.35$) while keeping tool use moderate ($\kappa{=}0.19$). \emph{nFrames Gating} alone pushes tool calls aggressively ($\kappa$ off-chart toward $1.36$) but leaves format stuck near $0.10$. The full recipe combines stronger format reward with moderate tool use: $f_\tau$ rises past step $45$ to a peak ($0.64$) that neither single component attains, while $\kappa$ stays at $0.21$.

\paragraph{Dispatch Mode.}
Block~C compares parallel and sequential dispatch on the same trained checkpoint. Parallel dispatch improves accuracy on every tested benchmark without retraining, with the largest gains on LongVideoBench and LVBench.

\section{Conclusion}
\label{sec:conclusion}

We present \textbf{\framework{}}, a multi-agent RL framework that dispatches parallel temporal crops in one turn for long-video reasoning, enabling peer-correctable evidence aggregation with fewer main-model generations. \method{} addresses the format instability and tool skipping observed in the Tool Prior Paradox by combining an anchor reward with per-prompt frame-budget randomization to make tool calls reward-relevant.
Supported by a self-curated $97$K-sample multi-task SFT corpus and a separate $4{,}406$-sample RL split spanning open-ended QA, multiple-choice, and temporal grounding, \framework{} outperforms existing open-source 7--8B baselines on six of seven long-video evaluation splits. These results support combining format anchoring with tool incentives on the tested backbone and motivate further study as tool capabilities become increasingly internalized in modern base LMMs~\citep{qwen2025qwen3vl,kimiteam2026kimik3}.

\begin{ack}
This work was supported by Apodex, which provided the compute, storage, and engineering infrastructure for the reported experiments. The authors declare no additional competing interests.
\end{ack}

\clearpage
\raggedbottom
\bibliography{camera_ready/paravt_cr}
\bibliographystyle{plainnat}

\clearpage
\appendix
\begin{center}
{\LARGE \textbf{Appendix}}
\end{center}
\vspace{0.5em}

\begin{itemize}[leftmargin=*]
    \item \textbf{Limitations and Broader Impact} (\Cref{app:limitations}): scope limits and dual-use considerations.
    \item \textbf{Implementation Details} (\Cref{app:impl}): hardware, SFT data composition and curation pipeline (sequential$\to$parallel conversion, Gemini-CoT distillation, format/storage), RL data and DAPO-inspired offline difficulty filtering, optimizer and reward coefficients, and token-budget accounting.
    \item \textbf{Prompts and Templates} (\Cref{app:prompts}): illustrative prompt structures, source-specific SFT layouts, and evaluation prompt classes.
    \item \textbf{Relation to Prior Work} (\Cref{app:prior_work}): component attribution and related reward designs.
    \item \textbf{Porting Checklist} (\Cref{app:porting}): adapting the recipe to another tool-native LMM.
    \item \textbf{Rollout Examples} (\Cref{app:rollouts}): three schematic examples illustrating format collapse and its mitigation under \method{}.
    \item \textbf{Cross-Model Evidence} (\Cref{app:obstacles}): per-tag format closure breakdown (\Cref{tab:format_breakdown}) and the two-model before/after trajectory (\Cref{fig:tool_prior_paradox}), extending \Cref{fig:cross_model}.
    \item \textbf{Tool Usage Patterns} (\Cref{app:tool_patterns}): training-time tool-call trajectories under three reward configurations (\Cref{fig:tool_usage}).
    \item \textbf{Negative Results} (\Cref{app:negative}): grouped by intervention axis: reward-shape (phase-reward staging, task-aware reward coefficients), data-shape (Pre-RFT, stronger cold-start), and gradient/format-shape (Token-Decoupled GRPO structural mask, bidirectional tag reversion).
\end{itemize}

\FloatBarrier
\section{Limitations, Broader Impact, and Future Work}
\label{app:limitations}

\paragraph{Limitations.}
\emph{(i)} The RL stage delivers its primary contribution as training-time format and tool-use stability under temperature sampling rather than as a large standalone greedy-eval delta on top of the cold-started checkpoint. Further amplifying the eval-time translation is an open direction. \emph{(ii)} The cross-model evidence for the role of the prior comes from a single Qwen2.5-VL vs.\ Qwen3-VL contrast (consistent with a causal interpretation but not equivalent to a controlled intervention). The full \method{} pipeline has only been validated on Qwen3-VL-8B. Its intended scope is LMMs that already carry a usable tool prior: on a weak-prior backbone the format machinery transfers but no tool behavior emerges for it to stabilize. Porting the recipe to other model families requires mapping their reasoning, answer, and tool-call tags (\Cref{app:porting}). A controlled sweep over prior strength across families is future work. \emph{(iii)} Only the \texttt{crop\_video} tool is evaluated. Whether the recipe generalizes to other tool families (text retrieval, scene-graph extraction, audio transcription) is open.

\paragraph{Broader Impact.}
\label{app:broader}
Agentic long-video understanding lowers the human cost of searching extended footage by content, with applications in accessibility, sports analytics, and archival retrieval. The same capability also reduces the marginal cost of large-scale surveillance over CCTV or body-camera streams, and \framework{}'s parallel-tool dispatch amplifies that throughput rather than restraining it. Deployment in such contexts should be paired with explicit consent and transparency frameworks. The \method{} training recipe is tool-agnostic and could be retargeted to tool families with different safety profiles (\eg document or person retrieval), so the dual-use surface is broader than the \texttt{crop\_video} tool we evaluate. We release code, data, and weights to enable independent audit but withhold surveillance-specific finetunes. Downstream users adapting \method{} to higher-risk tool families should conduct their own impact assessment.

\paragraph{Future Work.}
\emph{(i)}~Scaling \method{} to larger LMMs ($32$B--$72$B) where richer base capabilities may make RL exploration more effective. \emph{(ii)}~Extending necessity gating to other agentic settings where tool necessity is not guaranteed, such as retrieval-augmented generation and code execution.
\FloatBarrier
\section{Implementation Details}
\label{app:impl}

\paragraph{Hardware.}
All experiments are conducted on 2 machines, each with 8$\times$ NVIDIA GPUs ($\geq$80\,GB VRAM each).
For RL training, we use 7 GPUs for FSDP parameter updates and 1 GPU for SGLang inference serving.

\paragraph{Base Model.}
We use Qwen3-VL-8B-Instruct~\citep{qwen2025qwen3vl} as the base LMM.
Training overview processing is configured at $1$\,fps with a $64$-frame cap. The released evaluation driver instead samples $64$ overview frames uniformly. Temporal crops use $1$\,fps sampling with at most $16$ frames.

\paragraph{Training Infrastructure.}
The AReaL framework pipelines rollout generation (GPU 0, SGLang) with FSDP training (GPUs 1--7).
After the first step, rollout wait drops from ${\sim}500$s to ${<}2$s due to pipelining.
Each training step takes approximately 50 minutes, including ${\sim}35$ minutes for rollout and ${\sim}15$ minutes for parameter updates.
We set \texttt{SGLANG\_VLM\_CACHE\_SIZE\_MB=4096} to accommodate 64-frame video embeddings (${\sim}134$\,MB each).

\paragraph{Multi-Agent Rollouts and Loss Masking.}
In RL rollouts, every \texttt{<tool\_call>} parsed from a main-agent turn launches a sub-agent, a fresh single-turn call of the current policy. Its input holds only the frames of the requested window, decoded at $1$\,fps with at most $16$ frames of at most $224{\times}224$ pixels, and a fixed, query-agnostic prompt. The prompt asks for a concise summary of the segment. The calls of one turn run concurrently. Each reply is truncated to $512$ tokens and appended to the main agent's context as a user-role \texttt{<tool\_response>} turn. Sub-agent generations are excluded from the loss: the main-agent response receives a sequence-level scalar reward, and its group-normalized advantage weights the policy loss across all trained main-agent tokens. The optimized object is the main agent's dispatch and aggregation behavior, and sub-agents change only through the shared weights. In SFT, summaries are also supplied as user-role context, and multiple \texttt{<tool\_response>} blocks may share one user turn. Of the $96{,}768$ SFT samples, $29{,}122$ ($30.1\%$) contain \texttt{<tool\_response>} blocks, all of them in user-role turns. The SFT loss is computed on assistant turns only, so the summaries are consumed as context but never supervised.

\paragraph{Evaluation Harness.}
Tool-augmented models in \Cref{tab:main} are evaluated with greedy decoding in a multi-turn harness that allows at most $3$ tool turns and requests a final answer after the last one. Each crop returns its frames, decoded as above, directly to the calling model, and \framework{} may issue up to $5$ crops per turn. Each baseline keeps its own prompt class and tool schema (\Cref{app:prompts}).

\paragraph{SFT Configuration.}
SFT uses the lmms-engine framework with FSDP.
We train for 1,500 steps total, with checkpoints at every 100 steps.
Learning rate: $2 \times 10^{-5}$, batch size: 32, optimizer: AdamW.
The cold-started (step $500$) checkpoint is selected as the RL initialization based on training-time format stability under temperature sampling (\Cref{sec:ablation}).

\noindent\begin{minipage}{\linewidth}
\paragraph{SFT Data (97K samples).}
The SFT training set contains 97K samples from 7 sources:

\begin{table}[H]
\caption{\textbf{SFT data composition.} ``Tool'' indicates whether the sample contains parallel \texttt{crop\_video} calls.}
\label{tab:sft_data}
\small
\centering
\resizebox{\linewidth}{!}{%
\begin{tabular}{l r l l}
\toprule[1pt]
\textbf{Dataset} & \textbf{Samples} & \textbf{Tool?} & \textbf{Description} \\
\midrule
LLaVA-Video-178K (subsampled) & 50K & No & General video QA from the VideoR1 training pool \\
Self-trace (LongVT rollouts) & 15K & Yes & Sequential$\to$parallel converted tool traces \\
Charades-STA & 12K & No & Temporal grounding (\texttt{[start, end]}) \\
TVG (Charades-STA) & 6K & Yes & Temporal grounding with parallel tool calls \\
Long video reasoning & 5K & No & Multi-step reasoning on 5--60 min videos \\
Gemini-CoT (distilled) & 5K & Yes & Tool traces generated by Gemini-2.5-Flash \\
MUSEG (multi-segment) & 2.5K & Yes & Avg 4.4 parallel \texttt{crop\_video} calls/sample \\
\bottomrule[1pt]
\end{tabular}%
}
\end{table}

\end{minipage}\par

\paragraph{Tool-Augmented Fraction.}
\label{app:tool_fraction}
Tool-bearing samples comprise $30\%$ of the SFT mix. This fraction was set by a Plan~A vs.\ Plan~B comparison: an earlier Plan~A ($212$K total, $14\%$ tool) produced weaker tool-calling behavior in downstream RL than the current Plan~B ($97$K total, $30\%$ tool), despite having more raw samples. We read this as evidence that the fraction of tool-bearing samples matters more than absolute count once a non-trivial volume of non-tool video QA is present, and we did not re-tune the ratio further.

\paragraph{Sequential$\to$Parallel Conversion.}
The selftrace, Gemini-CoT, and TVG sources start as sequential single-tool LongVT-style traces (one \texttt{<tool\_call>} per assistant turn, with cropped frames re-injected into the next turn's context). We convert each trace to a single-turn parallel format by merging consecutive \emph{independent} tool calls into one turn. We treat two adjacent calls as independent when their target windows do not overlap and the tool responses they consumed contain no cross-reference to one another (typical case: ``inspect 00:30--00:50'' followed by ``inspect 02:10--02:25,'' both grounded in disjoint visual evidence). Calls that fail this check (for example, a follow-up crop refining the timestamps of a previous response) are kept on their own turn. We then replace each tool's frame response with a textual summary of the segment's visual content, drawn from the LongVT model's existing assistant continuation that consumed those frames. The text-summary substitution serves two purposes: it aligns the SFT data with the RL sub-agent's output format (text, not frames) and it keeps context length manageable when several crops appear in the same response. After conversion, MUSEG remains the only source with consistently many parallel calls per turn ($\sim 4.4$ on average). Other sources average close to one call per turn because most LongVT traces issued only one crop to begin with.

\paragraph{Gemini-CoT Distillation.}
The $5$K Gemini-CoT subset is produced by sampling LongVT-selfQA prompts and generating sequential tool traces with Gemini-$2.5$-Flash, then running them through the same sequential$\to$parallel conversion above. Two practical issues drove additional steps. First, Gemini's content filter refuses certain video-question pairs. For those pairs, we re-issue the prompt to Qwen3-VL-$235$B as a fallback distiller and accept its trace if it passes downstream validation. Second, raw model outputs occasionally contain JSON-structural noise (unbalanced braces, prose around the tool call). We run a GPT-$4$o cleanup pass that re-emits each tool call as a strict JSON block and discards any sample violating \texttt{start\_time}$\,<\,$\texttt{end\_time} or with an empty answer field.

\paragraph{Format and Storage.}
All splits are stored as Parquet files with the \texttt{messages} column serialized as a JSON string, sidestepping Arrow's schema requirement when messages have heterogeneous tool-call structure. Chat layouts vary by source: direct Charades-STA examples contain a user query and an assistant answer, whereas tool traces include additional tool-response context. The configured video parameters are \texttt{max\_pixels}\,$=$\,$50176$, \texttt{fps}\,$=$\,$1$, and \texttt{max\_frames}\,$=$\,$64$.

\paragraph{RL Data ($4{,}406$ samples).}
The RL training set aggregates three task families: $1{,}606$ open-ended QA from filtered LongVT-selfQA-v$2$ (HACS / Ego$4$D-NaQ source videos), $1{,}600$ multiple-choice from the VideoR1 pool, and $1{,}200$ temporal-grounding queries from the Charades-STA training split. SFT and RL share some Charades training videos; the Charades test split is held out for evaluation. The open-ended QA pool starts from $1{,}668$ raw samples. An offline difficulty filter inspired by DAPO~\citep{yu2026dapo} excludes $55$ prompts with ground-truth answers longer than $15$ words and $7$ prompts with unanimously unsuccessful cold-start rollouts. These exclusions are heuristics based on the cold-started policy, not guarantees of zero advantage under later policies. The filter runs once before RL.

\noindent\begin{minipage}{\linewidth}
\paragraph{RL Configuration.}
GRPO training uses the AReaL asynchronous RL framework with the following hyperparameters:

\begin{table}[H]
\small
\centering
\caption{\textbf{RL training hyperparameters.}}
\label{tab:hyperparams}
\begin{tabular}{ll}
\toprule[1pt]
\textbf{Parameter} & \textbf{Value} \\
\midrule
Learning rate & $2 \times 10^{-6}$ \\
Warmup & 0 steps \\
Temperature ($\tau$) & 0.7 \\
GRPO group size ($G$) & 8 \\
Batch size & 7 (must be divisible by the number of FSDP workers (7)) \\
Max new tokens & 2048 \\
KL coefficient & 0.01 \\
Clip ratio ($\epsilon$) & 0.2 \\
Reward bias & $-0.2$ \\
FORMAT\_WEIGHT & 1.0 \\
ANCHOR\_WEIGHT & 0.5 \\
nFrames Gating set & \{4, 8, 16, 32, 64\} \\
Anchoring $\alpha, \beta, \gamma$ & 0.4, 0.3, 0.3 \\
Necessity bonus & disabled (ablation: negligible effect) \\
Contrastive reward & disabled (ablation: hurt performance) \\
Phase reward staging & disabled (ablation: 160 steps, no improvement) \\
\midrule
\multicolumn{2}{l}{\emph{Infrastructure}} \\
\midrule
Framework & AReaL (async RL, FSDP + SGLang) \\
GPU allocation & 1 GPU (SGLang inference) + 7 GPUs (FSDP training) \\
SGLang VLM cache & 4096\,MB (\texttt{SGLANG\_VLM\_CACHE\_SIZE\_MB}) \\
SGLang watchdog timeout & 600\,s \\
Checkpoint save frequency & every 5 steps \\
\bottomrule[1pt]
\end{tabular}
\end{table}

\end{minipage}\par

\paragraph{Reward Function Details.}
Instantiating \Cref{eq:reward} with the released defaults ($\lambda_{\text{fmt}}{=}1.0$, $\lambda_{\text{anchor}}{=}0.5$) gives:
\begin{equation*}
R(x, y) = R_{\text{acc}}(y, a^*) + R_{\text{fmt}}(y) + R_{\text{tool}}(y), \qquad R_{\text{fmt}}(y) = R_{\text{base}}(y) + 0.5 \cdot R_{\text{anchor}}(y).
\end{equation*}
Under the released default (\texttt{ANSWER\_SUFFIX} on), the base format reward $R_{\text{base}}$ assigns partial credit: $+0.2$ for substantive \texttt{<think>} content ($\geq$10 chars), $+0.3$ for \texttt{<answer>} tag, $+0.2$ for \texttt{</answer>} tag, $+0.3$ for correct think$\to$tool ordering, and $+0.1$ for balanced tag pairs.
The anchoring component $R_{\text{anchor}}$ is defined in \Cref{eq:anchor} with $(\alpha, \beta, \gamma){=}(0.4, 0.3, 0.3)$.
This sequence-level scalar contributes to the total reward, whose group-normalized advantage is applied across all trained main-agent tokens. The format reward is a shaped score, not a binary parseability rate.

The answer extraction follows a 3-level fallback: (1) content within \texttt{<answer>} tags; (2) if no \texttt{<answer>} tag, content after \texttt{</think>} excluding tool calls; (3) last non-empty line.
Early detection of degenerate outputs (responses containing 5+ \texttt{<|im\_start|>} tokens in under 300 characters) short-circuits to zero reward.

\noindent\begin{minipage}{\linewidth}
\paragraph{Longer Training and Larger Batches.}
\Cref{tab:train_scale} reports the extended-training and larger-batch controls.

\begin{table}[H]
\centering

\caption{\textbf{RL Training Scale.} Extended training and a larger batch size from the cold-started checkpoint.}
\label{tab:train_scale}
\small
\resizebox{\linewidth}{!}{%
\begin{tabular}{l c c c c c c}
\toprule
\textbf{Setting} & \textbf{VMME w/o} & \textbf{VMME w/} & \textbf{LongVB} & \textbf{LVBench} & \textbf{MLVU} & \textbf{MMVU} \\
\midrule
Released checkpoint & $62.1$ & $69.4$ & $60.4$ & $39.8$ & $65.0$ & $68.6$ \\
Batch $7$, $100$ steps & $62.8$ & $70.2$ & $61.1$ & $39.1$ & $65.5$ & $67.7$ \\
Batch $7$, $200$ steps & $62.9$ & $69.8$ & $60.9$ & $39.0$ & $65.3$ & $67.5$ \\
Batch $7$, $300$ steps & $62.5$ & $70.9$ & $61.8$ & $41.2$ & $65.6$ & $69.6$ \\
Batch $14$ & $62.3$ & $69.2$ & $61.8$ & $40.4$ & $65.2$ & $69.5$ \\
\bottomrule
\end{tabular}%
}

\end{table}

\end{minipage}\par

\noindent\begin{minipage}{\linewidth}
\paragraph{Token-Budget Accounting: Parallel vs.\ Sequential.}
We distinguish main-agent input-token accounting from end-to-end latency. Consider one batch of $K$ independent crops, with an initial prompt of $P$ tokens and $C$ returned tokens per crop. If each request reprocesses its full context, sequential dispatch requires $K$ request generations followed by a final answer, whereas parallel dispatch requests all crops together and then generates an answer. Ignoring the generated request text, their main-agent input-token totals are approximately
\begin{equation}
T_{\text{seq}}(K) \approx (K+1)P + \frac{K(K+1)}{2}C,
\qquad
T_{\text{par}}(K) \approx 2P + KC.
\end{equation}
Here $C$ can represent text summaries during RL or visual tokens in the frame-return evaluation harness. Visual-token counts depend on the processor's spatial and temporal grids, rather than a fixed number per frame. This accounting excludes sub-agent computation and cache reuse; measured latency and memory are reported separately in \Cref{tab:efficiency}.
\end{minipage}\par

\FloatBarrier
\section{Prompts and Templates}
\label{app:prompts}

The templates below illustrate the reasoning and tool-call structure used in the pipeline. Exact wording and chat layouts can vary by data source and stage.

\paragraph{SFT cold-start system prompt (tool-augmented sources).}
The following illustrates the tool-augmented prompt structure:

\begin{verbatim}
You are a video understanding agent.

# Workflow
1. Think inside <think>...</think> about which video
   segments contain the evidence needed to answer.
2. Call tools using <tool_call>...</tool_call> blocks.
   You may issue multiple <tool_call> blocks in one turn
   to inspect different temporal windows in parallel.
3. After receiving <tool_response>, place your final
   answer inside <answer>...</answer>.

# Format
<think>your reasoning here</think>
<tool_call>{"name": "crop_video",
            "arguments": {"video_path": "...",
                          "start_time": ...,
                          "end_time": ...}}</tool_call>
... (more <tool_call> blocks if needed) ...
[After tool responses arrive]
<answer>your final answer</answer>

# Important
- ONLY use <tool_call> with the JSON format above.
- Do NOT use <tool_code>, Python syntax, or any other
  tool format.
- Do NOT call the same temporal window twice.
\end{verbatim}

\paragraph{SFT cold-start system prompt (non-tool sources).}
Non-tool examples use task-appropriate question-answering prompts. The following illustrates reasoning-style supervision; direct grounding examples can instead require only an answer block:

\begin{verbatim}
You are a helpful video understanding assistant.
First, reason about the question inside <think>...</think>.
Then provide your final answer inside <answer>...</answer>.
\end{verbatim}

\paragraph{Evaluation prompts (per baseline class).}
We select prompt classes according to model training paradigms (\Cref{sec:setup}). \emph{Direct-answer} (Qwen2.5-VL-Instruct): the question is asked verbatim, with no system prompt and no required output schema. \emph{Reasoning} (Video-R$1$, VideoRFT, VideoChat-R$1$, Video-Thinker, Time-R$1$, ReWatch-R$1$): a common \texttt{<think>...</think><answer>...</answer>} evaluation template. This does not imply that every reasoning model was trained on identical tags; ReWatch-R1, for example, uses action and observation markers in its original prompt~\citep{zhang2025rewatchr1}. \emph{Agentic} (Qwen3-VL, Conan, \framework{}): the parallel tool prompt above, which uses Qwen's native \texttt{<tool\_call>} JSON schema and allows multiple calls per turn. The main-agent/sub-agent role split follows the multi-agent-in-one-model setup of MATPO~\citep{mo2025matpo}, whose planner prompt instead uses a Model Context Protocol (MCP)-style schema with one tool call per message. The remaining tool-trained baselines use their published native schemas (\texttt{<video\_zoom>} for VideoZoomer~\citep{ding2025videozoomer}, the Context-VLM JSON for SAGE~\citep{jain2025sage}, the interleaved multimodal chain-of-tool-thought suffix for LongVT-RFT~\citep{yang2026longvt}).

\FloatBarrier
\noindent\begin{minipage}{\linewidth}
\section{Relation to Prior Work}
\label{app:prior_work}
\Cref{tab:attribution} lists the components of \method{} that come from prior work. Format rewards and the reasoning template are standard. \method{} combines an anchor reward with nFrames Gating. In our ablations, the combination improves format reward over either single-component variant while retaining tool use (\Cref{sec:ablation}).

\begin{table}[H]
\centering
\caption{\textbf{Components of \method{} and Their Prior Work.}}
\label{tab:attribution}
\small
\begin{tabular}{@{}p{0.19\linewidth}p{0.30\linewidth}p{0.44\linewidth}@{}}
\toprule[1pt]
\textbf{Prior work} & \textbf{What it contributes} & \textbf{Relation to \method{}} \\
\midrule
DeepSeek-R1~\citep{guo2025deepseek} & \texttt{<think>}/\texttt{<answer>} template; rule-based format reward & Adopted (Think Prefix; base format reward $R_{\text{base}}$) \\
Constrained decoding~\citep{willard2023efficient} & Grammar-constrained generation over the vocabulary & The Think Prefix is its simplest instance, a fixed prefix \\
Open-R1~\citep{openr12025} & Equal credit for each reasoning and answer tag & Closest prior to our anchor reward, which also penalizes unclosed reasoning blocks \\
Datarus-R1~\citep{benchaliah2025datarus} & Structural format reward combined with a semantic reward & Same class as $R_{\text{base}}$ in our vanilla GRPO baseline \\
DAPO~\citep{yu2026dapo} & Clip-Higher, dynamic sampling, token-level loss, overlong shaping & Our offline difficulty filter differs from its online sampling (\Cref{sec:data}) \\
\bottomrule[1pt]
\end{tabular}
\end{table}

\end{minipage}\par
\FloatBarrier

\FloatBarrier
\section{Porting Checklist}
\label{app:porting}
To apply \method{} to another tool-native LMM: \emph{(i)} identify the reasoning, answer, and tool-call tags of its chat template and set the Think Prefix to its reasoning opener; \emph{(ii)} point the closure checks of $R_{\text{anchor}}$ and the base format reward at those tags; \emph{(iii)} before RL, sample the cold-started policy at the training temperature and record its format reward and tool-call rate, the two quantities \method{} is meant to keep jointly non-trivial; \emph{(iv)} keep part of the nFrames Gating support below the full overview budget, scaled to the frame limit of the backbone. We regard a port as successful when, relative to vanilla GRPO on the same backbone, the format reward at least doubles, the tool-call rate stays at least half of its pre-RL value, and accuracy does not decrease.

\FloatBarrier
\section{Rollout Examples}
\label{app:rollouts}

The following schematic examples illustrate the intended tag structure, Format Fragility, and its mitigation. Tool calls are abbreviated as pseudocode; executable calls use the JSON schema documented above. These examples are illustrations rather than verbatim evaluation outputs.

\paragraph{Example 1: SFT Baseline (Greedy Decoding).}
The first example illustrates the intended structure under greedy decoding:
\begin{verbatim}
<think>The question asks about... I should examine
the segment from 01:20 to 02:30 more closely.</think>
<tool_call>crop_video("video.mp4", 80, 150)</tool_call>
<tool_call>crop_video("video.mp4", 200, 280)</tool_call>
[tool responses...]
<answer>The person picks up the cup at 01:45.</answer>
\end{verbatim}

\paragraph{Example 2: Early RL (Format Collapse).}
The second illustrates format collapse under temperature sampling ($\tau{=}0.7$) without \method{}:
\begin{verbatim}
<think>Looking at the video, I can see...
[reasoning continues for 500+ tokens]
[no </think> closure]
<tool_code>python
# Switches to an alternative tool format
crop_video("video.mp4", start=80, end=150)
</tool_code>
[no <answer> tag, output truncated]
\end{verbatim}

\paragraph{Example 3: \method{} (Recovered Format).}
The third illustrates the structure encouraged by \method{} under sampling:
\begin{verbatim}
<think>The question requires temporal localization.
With only 8 overview frames, I need to crop specific
segments for detail.</think>
<tool_call>crop_video("video.mp4", 75, 155)</tool_call>
<tool_call>crop_video("video.mp4", 195, 285)</tool_call>
[tool responses...]
<answer>The person picks up the cup around 01:42.</answer>
\end{verbatim}

\FloatBarrier
\section{Cross-Model Evidence}
\label{app:obstacles}

\noindent\begin{minipage}{\linewidth}
\subsection{Per-Tag Format Closure}

\Cref{tab:format_breakdown} reports the per-tag closure breakdown of the Format Fragility side of the paradox (\Cref{sec:intro}). Rows are computed from raw $\tau{=}0.7$ training-stream rollouts, so they reflect compliance \emph{during} RL exploration. Vanilla GRPO reduces the cold-start-learned closure rates within $9$ steps as the policy reward-hacks toward direct answering. \method{} restores all three rates above the cold-started (step $500$) baseline by step $19$.

\begin{table}[H]
\centering
\caption{\textbf{Format Fragility Quantified.} Each row reports the closure rates of the three structural tags in raw $\tau{=}0.7$ rollouts sampled from training streams. \texttt{<think>}: fraction of completions with a properly closed reasoning block. \texttt{<tool\_call>}: fraction with a closed and JSON-parseable tool-call block, the agentic tag whose collapse most directly disables tool-augmented reasoning. \texttt{<answer>}: fraction with a properly bracketed answer block. Vanilla GRPO reduces the cold-start-learned closure rates within $9$ steps as the policy reward-hacks toward direct answering. \method{} restores them via Selective Anchoring, a reward based on tag checks. $\dagger$ marks the full \method{} recipe.}
\label{tab:format_breakdown}
\small
\resizebox{\textwidth}{!}{%
\begin{tabular}{l c c c c}
\toprule[1pt]
\textbf{Configuration} ($\tau{=}0.7$ rollouts) & $N$ & \texttt{<think>} & \texttt{<tool\_call>} & \texttt{<answer>} \\
\midrule
Cold-start (step $500$) (pre-RL baseline) & 166 & 51.8\% & 47.4\% & 30.1\% \\
Phase~C step $9$ (vanilla GRPO, no \method{}) & 56 & 26.8\% & 18.2\% & 19.6\% \\
Constrained only step $9$ (no anchor) & 56 & 14.3\% & 9.5\% & 10.7\% \\
\method{} step $9$ (anchor active) & 56 & 30.4\% & 31.7\% & 28.6\% \\
\rowcolor{gray!15} \method{} step $19$ (anchor stabilized) $\dagger$ & 56 & \textbf{58.9\%} & \textbf{52.6\%} & \textbf{41.1\%} \\
\bottomrule[1pt]
\end{tabular}%
}
\end{table}

\end{minipage}\par

\noindent\begin{minipage}{\linewidth}
\subsection{Two-Model Trajectory}

\Cref{fig:cross_model} in the main body summarizes the cross-model contrast (Qwen2.5-VL vs.\ Qwen3-VL) at the two endpoints of the prior gradient. \Cref{fig:tool_prior_paradox} provides the complementary before/after view of the same two checkpoints over the full $520$-step training horizon.

\begin{figure}[H]
\centering
\includegraphics[width=\linewidth]{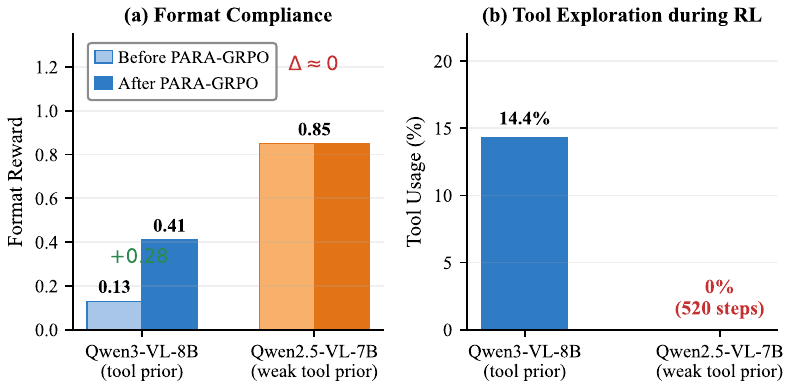}
\caption{\textbf{The Tool Prior Paradox (two-model trajectory).} \emph{(a)} Qwen3-VL's format climbs from $0.13$ to $0.41$ under \method{}. Qwen2.5-VL stays near $0.85$. \emph{(b)} Qwen3-VL keeps exploring tools ($14.4\%$ of rollouts contain a tool call). Qwen2.5-VL emits none over $520$ steps. Complements \Cref{fig:cross_model}.}
\label{fig:tool_prior_paradox}
\end{figure}
\end{minipage}\par

\FloatBarrier
\noindent\begin{minipage}{\linewidth}
\section{Tool Usage Patterns}
\label{app:tool_patterns}

\Cref{fig:tool_usage} traces training-time tool-call trajectories under three reward configurations (no penalty, no-tool penalty only, full \method{}).

\begin{figure}[H]
\centering
\includegraphics[width=\linewidth]{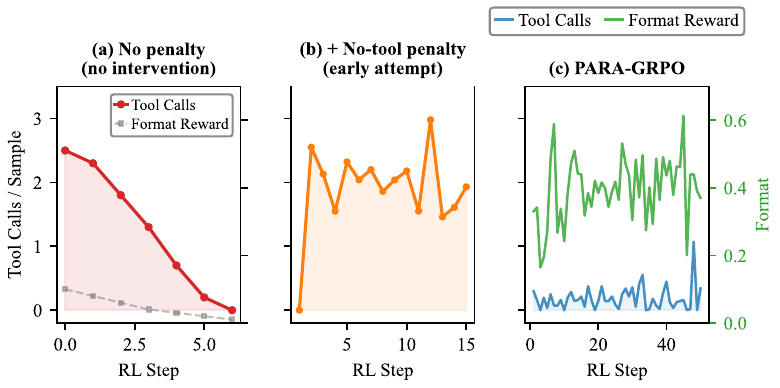}
\caption{\textbf{Training-time tool usage} (rollout averages at $\tau{=}0.7$, group size $8$). \emph{(a)} Early GRPO without intervention: tool usage drops from $2.5$ to $0$ by step $7$ (reward hacking). \emph{(b)} Adding a penalty for skipping tools keeps tool calls but format stays low. \emph{(c)} Under \method{}, $\kappa$ stabilizes at $0.1$--$0.5$ while $f_{\tau}$ (green, right axis) climbs to $0.41$.}
\label{fig:tool_usage}
\end{figure}
\end{minipage}\par

\FloatBarrier
\section{Negative Results and Failure Modes}
\label{app:negative}

We organize the negative results by the axis they intervene on: \emph{reward-shape} (Phase staging, Task-Aware coefficients), \emph{data-shape} (Pre-RFT, Stronger Cold-Start), and \emph{gradient/format-shape} (TD-GRPO mask, Bidirectional tag reversion). Each fails for a distinct reason that further constrains the design space.

\subsection{Reward-Shape Interventions}

\paragraph{Phase Reward Staging.}
We first optimized format reward in isolation, planning to introduce accuracy reward once format stabilized. After $160$ steps of format-only optimization, $f_{\tau}$ remained at $0.13$ with no upward trend, suggesting format and accuracy signals are interdependent: the model needs the accuracy gradient to motivate format learning in the first place.

\paragraph{Task-Aware Reward Coefficients.}
We added task-aware coefficients ($1.5{\times}$ for concise MCQ answers, $0.3$--$0.4{\times}$ for verbose ones) on top of \method{}. Training-time accuracy reward improves from $0.15$ to $0.24$ over $30$ steps and format reward stays comparable at $0.39$, but the variant does not outperform base \method{} on held-out eval: the best checkpoint reaches VideoMME $61.81$ (vs.\ \method{}'s $62.11$) and LongVideoBench $58.26$ (vs.\ $60.40$). Task-aware shaping improves training signal quality without translating into held-out eval gains, so we keep the simpler unweighted accuracy reward in the default recipe.

\subsection{Data-Shape Interventions}

\paragraph{Pre-RFT (rejection fine-tuning).}
We sampled $\tau{=}0.7$ rollouts from the cold-started checkpoint, filtered for format-compliant samples, and mixed them back into SFT training. Subsequent RL from this Pre-RFT init peaked at $f_\tau{=}0.40$, but this refinement degraded the cold-start quality on every downstream metric, so we did not adopt this pre-RL refinement route.

\paragraph{Stronger Cold-Start, Worse RL.}
Augmenting cold-start data with $12\%$ parallel tool-calling samples ($106$K vs.\ $97$K) produces a stronger cold-started checkpoint (VideoMME $61.3{\to}62.3$). RL from this stronger init produces zero tool calls throughout training. Three possible explanations are: \emph{(i)} the stronger model answers correctly without tools even under gating, so the tool-rewarded gradient is averaged out; \emph{(ii)} mixed single/parallel tool patterns in the cold-start data increase Format Fragility; \emph{(iii)} more thorough SFT coverage shifts the policy toward reproducing the cold-start distribution rather than exploring. These coupled effects motivate keeping the cold-start scope to the format schema rather than expanding it into the tool-call distribution itself.

\subsection{Gradient and Format-Shape Interventions}

\paragraph{Token-Decoupled GRPO (TD-GRPO) Structural Mask.}
We test a token-decoupled GRPO variant that selectively zeros the policy-gradient contribution of structural tokens (\eg \texttt{<think>}, \texttt{<tool\_call>}) so that RL only updates semantic content tokens, motivated by token-level analyses of sparse policy changes~\citep{meng2026sparse}. After $11$ steps, $f_{\tau}$ dropped to $0.11$ (\emph{below} baseline $0.13$): this masking did not preserve format in the tested run.

\noindent\begin{minipage}{\linewidth}
\paragraph{Bidirectional Format Reversion.}
\label{app:bidirectional}
Our main runs use SFT with \texttt{<tool\_call>}, which is Qwen3-VL's native tool-calling tag: it is present in the tokenizer vocabulary as a single added token (ID~151657) and is the format emitted by the model's default chat template.
To probe whether Format Fragility is competition between the SFT tag and inherited tool-format tendencies, we re-run SFT with \texttt{<tool\_code>} instead: a multi-subword tool-format sequence outside the tokenizer's added vocabulary.
The \texttt{<tool\_code>}-trained model still generates \texttt{<tool\_call>} in 5.4\% of rollouts (despite never seeing it during SFT), while its trained \texttt{<tool\_code>} appears in only 1.8\%.
This \emph{bidirectional format reversion} (\Cref{fig:bidir_reversion}) supports an interpretation of competing tool-format representations rather than a single mismatched tag: neither SFT target eliminates the alternative tag under temperature sampling. The probe does not identify the training stage at which these representations were acquired.
The format-substituted model also shows lower total tool emission ($7.2\%$ vs.\ $14.4\%$ of rollouts emit any tool tag), consistent with the probability-mass argument that single-token special tokens are more efficiently reinforced by RL than multi-subword sequences.

\begin{figure}[H]
\centering
\includegraphics[width=\linewidth]{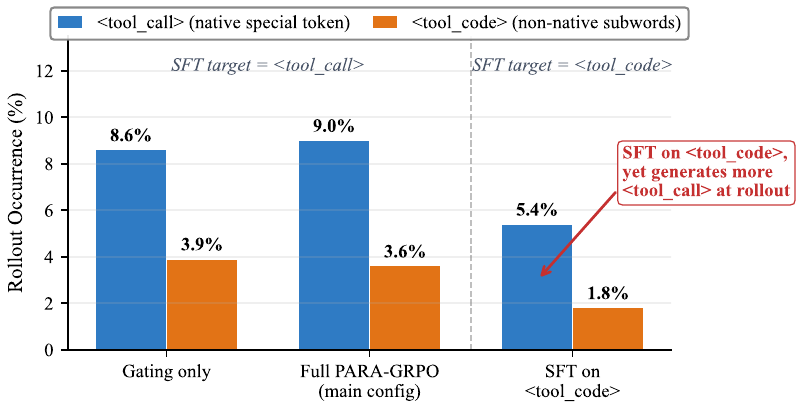}
\caption{\textbf{Bidirectional tag reversion during RL.} SFT with \texttt{<tool\_call>}: RL also emits \texttt{<tool\_code>} ($3.6$--$3.9\%$ of rollouts). SFT with \texttt{<tool\_code>}: RL still emits \texttt{<tool\_call>} more often ($5.4\%$) than the SFT-trained tag ($1.8\%$). Tag substitution does not remove Format Fragility.}
\label{fig:bidir_reversion}
\end{figure}
\end{minipage}\par

\clearpage
\section*{NeurIPS Paper Checklist}

\begin{enumerate}

\item {\bf Claims}
    \item[] Question: Do the main claims made in the abstract and introduction accurately reflect the paper's contributions and scope?
    \item[] Answer: \answerYes{}
    \item[] Justification: The abstract and \Cref{sec:intro} state the four contributions (parallel video tool architecture, identification of the Tool Prior Paradox, the \method{} framework, and the empirical study). Each is supported by results in \Cref{sec:exp} and the Appendix.

\item {\bf Limitations}
    \item[] Question: Does the paper discuss the limitations of the work performed by the authors?
    \item[] Answer: \answerYes{}
    \item[] Justification: \Cref{app:limitations} discusses single-model validation of the full \method{} pipeline (Qwen3-VL-8B only), its scope to backbones with a usable tool prior, single-tool scope (\texttt{crop\_video}), and the eval-deployment translation.

\item {\bf Theory assumptions and proofs}
    \item[] Question: For each theoretical result, does the paper provide the full set of assumptions and a complete (and correct) proof?
    \item[] Answer: \answerNA{}
    \item[] Justification: The paper is empirical and introduces no formal theorems or proofs. The reward and anchoring formulas in \Cref{sec:paragrpo} are definitional rather than theoretical claims.

\item {\bf Experimental result reproducibility}
    \item[] Question: Does the paper fully disclose all the information needed to reproduce the main experimental results of the paper to the extent that it affects the main claims and/or conclusions of the paper (regardless of whether the code and data are provided or not)?
    \item[] Answer: \answerYes{}
    \item[] Justification: \Cref{sec:setup} specifies the base model, SFT and RL data, training framework (AReaL), and 64-frame agentic eval protocol. \Cref{app:impl} lists the full SFT/RL hyperparameters, reward decomposition, infrastructure settings, and the SFT data composition (\Cref{tab:sft_data}).

\item {\bf Open access to data and code}
    \item[] Question: Does the paper provide open access to the data and code, with sufficient instructions to faithfully reproduce the main experimental results, as described in supplemental material?
    \item[] Answer: \answerYes{}
    \item[] Justification: Code (\url{https://github.com/EvolvingLMMs-Lab/ParaVT}, Apache-2.0), the SFT and RL data (\url{https://huggingface.co/datasets/ParaVT/ParaVT-Parquet} and \url{https://huggingface.co/datasets/ParaVT/ParaVT-Source}), and the \framework{}-8B weights (\url{https://huggingface.co/ParaVT/ParaVT-8B}) are publicly released with instructions to reproduce the main results.

\item {\bf Experimental setting/details}
    \item[] Question: Does the paper specify all the training and test details (e.g., data splits, hyperparameters, how they were chosen, type of optimizer) necessary to understand the results?
    \item[] Answer: \answerYes{}
    \item[] Justification: Training and evaluation details (data splits, optimizer, learning rates, batch size, GRPO group size, temperature, KL coefficient, frame count, eval protocol) are summarized in \Cref{sec:setup} and reported in full in \Cref{app:impl} (\Cref{tab:hyperparams}).

\item {\bf Experiment statistical significance}
    \item[] Question: Does the paper report error bars suitably and correctly defined or other appropriate information about the statistical significance of the experiments?
    \item[] Answer: \answerNo{}
    \item[] Justification: Multi-seed retraining of the full SFT$\to$RL pipeline was beyond our compute budget. We rely on training-time format and tool-use trajectories (\Cref{fig:training_curves,fig:tool_usage}) as the primary evidence for the RL contribution, alongside the held-out evaluation in \Cref{tab:main}.

\item {\bf Experiments compute resources}
    \item[] Question: For each experiment, does the paper provide sufficient information on the computer resources (type of compute workers, memory, time of execution) needed to reproduce the experiments?
    \item[] Answer: \answerYes{}
    \item[] Justification: \Cref{app:impl} reports the hardware (8 NVIDIA GPUs at $\geq$80\,GB VRAM each per RL run, with 7 FSDP + 1 SGLang, with evaluation on nodes of 4 to 8 such GPUs). It also reports the AReaL framework setup and $\sim\!50$ minutes per RL training step.

\item {\bf Code of ethics}
    \item[] Question: Does the research conducted in the paper conform, in every respect, with the NeurIPS Code of Ethics \url{https://neurips.cc/public/EthicsGuidelines}?
    \item[] Answer: \answerYes{}
    \item[] Justification: The work uses public benchmarks and a public base LMM under their original research licenses, and involves no human subjects or private user data. We have reviewed and complied with the NeurIPS Code of Ethics.

\item {\bf Broader impacts}
    \item[] Question: Does the paper discuss both potential positive societal impacts and negative societal impacts of the work performed?
    \item[] Answer: \answerYes{}
    \item[] Justification: A Broader Impact paragraph is included in the appendix (\Cref{app:broader}).

\item {\bf Safeguards}
    \item[] Question: Does the paper describe safeguards that have been put in place for responsible release of data or models that have a high risk for misuse (e.g., pre-trained language models, image generators, or scraped datasets)?
    \item[] Answer: \answerNA{}
    \item[] Justification: \framework{} is a long-video question-answering model fine-tuned on public benchmark data. The released assets are a video-QA fine-tune and annotations derived from public research datasets. Potential dual-use considerations are discussed in \Cref{app:broader}.

\item {\bf Licenses for existing assets}
    \item[] Question: Are the creators or original owners of assets (e.g., code, data, models), used in the paper, properly credited and are the license and terms of use explicitly mentioned and properly respected?
    \item[] Answer: \answerYes{}
    \item[] Justification: All used assets are cited at first mention.

\item {\bf New assets}
    \item[] Question: Are new assets introduced in the paper well documented and is the documentation provided alongside the assets?
    \item[] Answer: \answerYes{}
    \item[] Justification: The \framework{}-8B weights, the SFT and RL data, and the code are released with documentation (README and model and dataset cards) at the URLs given in the answer on open access. The code is licensed under Apache-2.0.

\item {\bf Crowdsourcing and research with human subjects}
    \item[] Question: For crowdsourcing experiments and research with human subjects, does the paper include the full text of instructions given to participants and screenshots, if applicable, as well as details about compensation (if any)?
    \item[] Answer: \answerNA{}
    \item[] Justification: The paper does not involve crowdsourcing or research with human subjects. All training and evaluation data come from existing public datasets.

\item {\bf Institutional review board (IRB) approvals or equivalent for research with human subjects}
    \item[] Question: Does the paper describe potential risks incurred by study participants, whether such risks were disclosed to the subjects, and whether Institutional Review Board (IRB) approvals (or an equivalent approval/review based on the requirements of your country or institution) were obtained?
    \item[] Answer: \answerNA{}
    \item[] Justification: No human-subjects research is conducted in this paper, so IRB review is not applicable.

\item {\bf Declaration of LLM usage}
    \item[] Question: Does the paper describe the usage of LLMs if it is an important, original, or non-standard component of the core methods in this research? Note that if the LLM is used only for writing, editing, or formatting purposes and does \emph{not} impact the core methodology, scientific rigor, or originality of the research, declaration is not required.
    \item[] Answer: \answerYes{}
    \item[] Justification: A large multimodal model (Qwen3-VL-8B) is the research subject of this paper and is the artifact being trained, as detailed in \Cref{app:impl}. Gemini-2.5-Flash was additionally used to generate 5K tool-calling traces in the SFT data (\Cref{tab:sft_data}), with Qwen3-VL-235B as a fallback distiller and GPT-4o for a JSON cleanup pass (\Cref{app:impl}).

\end{enumerate}

\end{document}